\documentclass[times,twocolumn,final,authoryear]{elsarticle}

\usepackage{framed,multirow}
\usepackage{graphicx} 
\usepackage{subcaption}

\newsavebox{\twosubbox}
\usepackage{amsmath} 
\usepackage{amssymb}  
\usepackage{algorithm}
\usepackage[noend]{algpseudocode}
\usepackage{bm}
\usepackage{latexsym}

\usepackage{url}
\usepackage{xcolor}
\definecolor{newcolor}{rgb}{.8,.349,.1}

\usepackage[final]{review} 
\setrevision{1} 

\begin{document}

\begin{frontmatter}

\title{Plane-extraction from depth-data using a Gaussian mixture regression model
}
\author[1]{Richard T. Marriott}
\author[1]{Alexander Pashevich}
\author[1]{Radu Horaud} 

\address[1]{INRIA Grenoble Rh\^one-Alpes \& Univ. Grenoble Alpes, France}


\begin{abstract}
We propose a novel algorithm for unsupervised extraction of piecewise planar models from depth-data. Among other applications, such models are a good way of enabling autonomous agents (robots, cars, drones, etc.) to effectively perceive their surroundings and to navigate in three dimensions. 
We propose to do this by fitting the data with a piecewise-linear Gaussian mixture regression model whose components are skewed over planes, making them flat in appearance rather than being ellipsoidal, by embedding an outlier-trimming process that is formally incorporated into the proposed expectation-maximization algorithm, and by selectively fusing contiguous, coplanar components. Part of our motivation is an attempt to estimate more accurate plane-extraction by allowing each model component  to make use of all available data through probabilistic clustering. The algorithm is thoroughly evaluated against a standard benchmark and is shown to rank among the best of the existing state-of-the-art methods.
\end{abstract}



\end{frontmatter}



\section{Introduction}

The objective of this paper is to construct simple planar models of environments by identifying flat surfaces within depth-data. We propose to do this by (i) fitting the data with a piecewise-linear Gaussian mixture regression (GMR) model -- a Gaussian mixture model (GMM) whose components are \textit{skewed} over planes, making them flat in appearance rather than being ellipsoidal; and then (ii) selectively \textit{fusing} contiguous, coplanar components. Part of our motivation for evaluating this method was to attempt to estimate more accurate model parameters by allowing each model component to make use of all available data through probabilistic clustering. This contrasts with most other recent methods \citep{c4_8}, \citep{c7_11}, \citep{c5_19}, \citep{c3_20}, \citep{c6_23}, \citep{c9_29} which, for the sake of efficiency, compromise by working with noisier subsets of data-points.
The application in which we are specifically interested is the perception of a 3D environment by a non-human observer in order to enable navigation within that environment. The observer may be a wheeled or a legged robot, a drone, a driver-less car, a human perception-aid such as that seen in \citep{c1}, or any other similar device. 

Recently, dense depth-data have become readily available due to the development of affordable structured light and time-of-flight cameras. Each of these sensor-types produces images of depth-related values that can be projected as clouds of 3D points. These point-clouds, however, are nothing more than a noisy set of points that only sample the environment. The observer must then be able to make sense of these observations by using them to construct a model of some form, e.g. a set of planar surfaces.

An alternative to a piecewise-planar model might be to attempt to represent the environment as a set of known objects. To do so, however, comprehensive object-recognition training would be required.
In practice, in a dynamic, real-world environment, such a technique would ultimately only be able to complement a more general, unsupervised approach. Planar primitives are sufficiently general to model most environments. They are particularly appropriate in the home and office, where planar surfaces are prevalent, but can also handle more complex scenes, approximating curved surfaces in a piecewise fashion.
 Although a piecewise planar representation of the environment may not allow many objects to be identified, it provides a certain set of very useful semantics. Namely, the observer knows that it can navigate safely on roughly horizontal planes and that it cannot pass through roughly vertical ones.

The main contribution of our paper is a probabilistic treatment of the problem of extracting planes from depth images. We propose to combine piecewise linear regression with GMM \citep{deleforge2015high},  thus yielding an expectation-maximization (EM) algorithm, with proven mathematical convergence, that deterministically clusters the 3-D data into 2-D Gaussian components via likelihood maximization. Moreover, we use a recently proposed trimming method \citep{c12} that, unlike random sampling such as RANSAC-based methods, can be embedded within EM in a principled way. We demonstrate, using a standard benchmark, that accuracy of depth-image segmentation by our robust GMR technique is comparable with the best of the other state-of-the-art methods.\footnote{Supplemental material can be found at \url{https://team.inria.fr/perception/research/plane-extraction/}.}

%

\section{Related work}\label{sec:related_work}

There are many different methods of plane-extraction. These methods tend not to rely on single concepts but, instead, combine various \textit{component-algorithms} in different ways. There are three components that are typically used: \addnote[reorder]{1}{(i) Region-growing, whether it be to grow regions pixel by pixel or to absorb some form of nearby superpixels; (ii) Pixel-clustering; and (iii) RANSAC plane-fitting, usually applied to local regions only \citep{c4_8} \citep{c6_23} \citep{c9_29}}.

In \citep{c7_11} and \citep{c5_19}, various region-growing concepts are used. E.g. \citep{c5_19} performs per-pixel region-growing based on per-point normal-orientation and combined mean squared error (MSE). A second, larger-scale merging of regions is then performed to collect together planes that may have become disjoint due to noise in the original surface-normals. In \citep{c7_11}, some of the noise of per-point normal-estimation is reduced by first creating a grid of superpixels organised in an adjacency graph. Agglomerative hierarchical clustering\footnote{Despite it's name, the AHC in \citep{c7_11} is actually performing region-growing on a set of superpixels due to the restriction of the adjacency graph.} (AHC) is then used to merge the superpixels followed by per-pixel region-growing to refine the \textit{sawtooth} edges caused by the initial grid.

There are many examples of algorithms that perform clustering. In \citep{c3_20}, per-pixel normal-estimation is performed and then clustering by discrete values of normal-orientation and of perpendicular distance to the origin. Further pixel-by-pixel refinement is then performed to capture those points falling just on the wrong side of the discretization boundaries from the value of a dominant plane. In \citep{c4_8}  the gradient of depth (GoD) features are clustered: Points belonging to the same plane will have the same GoD across them. Once clusters are found, RANSAC plane-fitting is applied followed by merging of nearby planes. In \citep{pham2016geometrically} an adjacency graph is constructed over local surface patches and a graph clustering algorithm is then applied. Plane extraction is formulated as the minimization of a global pairwise energy function which jointly considers plane fidelities and geometric consistencies between planes, i.e. orthogonal or parallel planes.

A standard plane-extraction approach is to run RANSAC sequentially until no more planes can be found \citep{c10_16}. \citep{c6_23} and \citep{c9_29} use RANSAC for robust plane-fitting, applying it to local regions only, for efficiency. Clusters belonging to the resulting planar components are then grown to include surrounding points. \citep{c9_29} finds the initial local regions via a Hough transform-based pre-segmentation. In \citep{gallo2011cc} RANSAC is applied to connected components of inliers. In \citep{qian2014ncc} a coherence check is performed to remove data patches whose normals are in contradiction to the fitted planes, followed by a recursive plane-clustering process. One drawback of RANSAC-based methods is that they do not consider fusion of planar sets of points and hence they often under-estimate the number of actual planes.

In this work, we introduce the \textit{robust piecewise-linear} Gaussian mixture regression (RPL-GMR) algorithm for optimally fitting a set of planes to a 3D point cloud. The algorithm contains an outlier-trimming process, thus being able to replace RANSAC. In the literature, there are very few examples of using mixture models for plane-extraction. One example is \citep{c11}. Note, however, that the model used in \citep{c11} is a mixture of \textit{unbounded} planes that extend throughout the whole data-set. The idea of plane-locality, which is essential for good performance in more complex environments, is only introduced as a post-processing step. The RPL-GMR formulation is such that the locality of planes is estimated simultaneously with the planar parameters, making RPL-GMR a more powerful and elegant alternative to existing methods.

The rest of the paper is organised as follows: Section \ref{sec:GLLiM_theory} gives the RPL-GMR formulation and its associated EM algorithm; Section \ref{sec:robust_implementation} contains details of the various stages of the algorithm; in Section \ref{sec:benchmarking} our algorithm is evaluated against various others using the \textit{SegComp} data-set \citep{c18}; and in Section \ref{sec:conclusions} we draw conclusions.

\section{Piecewise-linear Gaussian mixture regression}\label{sec:GLLiM_theory}


The proposed model is a form of constrained GMM to find planar patches within sets of 3D data-points. A standard GMM would not be particularly useful and would find ellipsoid-like densities in the data. The model of \citep{deleforge2015high}, on the other hand, makes the assumption that data in high-dimensional space lie on a lower-dimensional manifold (corrupted only by uncorrelated Gaussian noise), and furthermore, that the surface can be well-approximated by a patchwork of locally linear functions. A model that makes these assumptions is ideal in our case where we have data-points measured at the 2D manifold which is the \textit{visible frontier} of the scene, and where we have scenes containing many planes, i.e. locally linear functions in the manifold.

Let this manifold be described by a function \begin{math}g: \mathcal{X}\mapsto \mathcal{Y}\end{math} where $\mathcal{X}\subset \mathbb{R}^2$ and $\mathcal{Y} \subset \mathbb{R}$. Obviously, \begin{math}g(\mathcal{X})\end{math} is not necessarily linear, in our case being composed of surfaces with various characteristics. Let $\mathbf{x}\in\mathcal{X}$ and $y\in\mathcal{Y}$ be realisations of the random variables $\mathbf{X}\in\mathbb{R}^2$ and $Y\in\mathbb{R}$.
The proposed model approximates the potentially nonlinear \begin{math}g(\mathbf{x})\end{math} in a piecewise linear fashion. As is common practice in mixture models, a discrete, hidden variable, \begin{math}Z \in \mathbb{N}\end{math} is introduced. The complete data then become \begin{math}(\bm{\mathbf{X}}, Y, Z)\end{math} where a realisation \begin{math}(\mathbf{x},y,Z=k)\end{math} of \begin{math}(\bm{\mathbf{X}},Y,Z)\end{math} indicates that \begin{math}y\end{math} is related to \begin{math}\mathbf{x}\end{math} by an affine mapping indexed by $k$, plus some error term, $e_k$. We assume, then, that \begin{math}g(\mathbf{x})\end{math} can be approximated by the following mixture of $K$ affine transformations:
\begin{align}
\label{eq:mapping}
Y = \sum_{k=1}^{K} \mathbb{I}(Z=k)(\mathbf{A_{k}\bm{\mathbf{X}}} + b_{k} + e_{k}),
\end{align}
where \begin{math}\mathbb{I}\end{math} is an indicator function such that \begin{math}\mathbb{I}(Z) = 1\end{math} if \begin{math}Z = k\end{math}, or \begin{math}0\end{math} otherwise; 
\addnote[rowvector]{1}{
$\mathbf{A}_k \in \mathbb{R}^{1\times2}$ 
}
and $b_k \in \mathbb{R}$ are the mapping parameters of the $k$-th affine transformation; and \begin{math}e_{k} \sim \mathcal{N}(0,\sigma_k), \sigma_k \in \mathbb{R}\end{math} is an error term capturing inaccuracies in both the observations and the mapping. Let the joint variable \begin{math}(\mathbf{X}, Y)\end{math} be modeled by a GMM:
\begin{align}
\label{eq:joint_var_gaussian}
p(\mathbf{x},y;\theta) = \sum_{k=1}^{K} \pi_{k} \mathcal{N}(\mathbf{x},y;\mathbf{m}_k, \mathbf{V}_k),
\end{align}
where \begin{math}\pi_k, \mathbf{m}_k\end{math} and \begin{math}\mathbf{V}_k\end{math} are the priors, means and covariances of the mixture, respectively. This is equivalent to:
\begin{align}
\label{eq:joint_prob_decomp_2}
p(\mathbf{x},y;\theta) = \sum_{k=1}^{K} p(y|\mathbf{x},Z=k;\theta)p(\mathbf{x}|Z=k;\theta)p(Z=k;\theta).
\end{align}
These probability distributions can be modeled as Gaussians, and so we have:
\begin{align}
\label{eq:marginal}
p(y|\mathbf{x},Z=k;\theta) &= \mathcal{N}(y;\mathbf{A}_k\mathbf{x} + b_k, \sigma_k), \\
\label{eq:locality}
p(\mathbf{x}|Z=k;\theta) &= \mathcal{N}(\mathbf{x};\mathbf{c}_k, \bm{\Gamma}_k),\\
\label{eq:pi}
p(Z=k;\theta) &= \pi_k,
\end{align}
where $\mathbf{c}_k \in \mathbb{R}^2$ and $\bm{\Gamma} \in \mathbb{R}^{2 \times 2}$ are, respectively, the centre and covariance of the Gaussian components in the space of $\bm{\mathcal{X}}$. Combining (\ref{eq:joint_prob_decomp_2}), (\ref{eq:marginal}), (\ref{eq:locality}) and (\ref{eq:pi}), we get the explicit expression for the joint probability of the observed data
\begin{align}
\label{eq:explicit_decomposition}
p(\mathbf{x},y;\theta) = \sum_{k=1}^{K} \pi_{k} \mathcal{N}(y;\mathbf{A}_k\mathbf{x} + b_k, \sigma_k) \mathcal{N}(\mathbf{x};\mathbf{c}_k, \bm{\Gamma}_k).
\end{align}
This is equivalent to the Gaussian distribution of the joint variable \begin{math}(\mathbf{X},Y)\end{math} in equation (\ref{eq:joint_var_gaussian}) where the mean vector and covariance matrix are given by
\begin{align}
\label{eq:joint_mean_and_cov}
\mathbf{m}_k = \left( \begin{array}{c} \mathbf{c}_k \\ \mathbf{A}_k\mathbf{c}_k + b_k \end{array} \right),
\mathbf{V}_k = \left( \begin{array}{cc} \bm{\Gamma}_k & \bm{\Gamma}_k \mathbf{A}_{k}^{\top} \\
\mathbf{A}_k \bm{\Gamma}_k & \sigma_k + \mathbf{A}_k \bm{\Gamma}_k \mathbf{A}_{k}^{\top} \end{array} \right).
\end{align}
The parameter set is
\begin{math}
\theta = \left\{\mathbf{c}_k, \bm{\Gamma}_k, \mathbf{A}_k, b_k, \sigma_k, \pi_k \right\}_{k=1}^{K}.
\end{math}


The RPL-GMR algorithm is an EM procedure that iteratively maximises the expectation of the complete-data log-likelihood with respect to the probability distribution of the hidden variables given the current model parameters:
\begin{align}
\label{eq:Q_function}
\mathcal{L}(\theta) = \sum_{k=1}^{K} \frac{1}{r_{k}} \sum_{n=1}^{N} r_{nk} \log(p(\mathbf{x}_n,y_n,Z_n=k;\theta)),
\end{align}
\addnote[numberpoints]{1}{
where $N$ is the number of data points}, $r_k=\sum_{n=1}^{N}r_{nk}$ and  \begin{math}r_{nk}\end{math} are the \textit{responsibilities}:
\begin{align}
\label{eq:explicit_responsibilities}
r_{nk} = \frac{\pi_{k} \mathcal{N}(y_n;\mathbf{A}_k\mathbf{x}_n + b_k, \sigma_k) \mathcal{N}(\mathbf{x}_n;\mathbf{c}_k, \bm{\Gamma}_k)}
{\sum_{i=1}^{K} \pi_{i} \mathcal{N}(y_n;\mathbf{A}_i\mathbf{x}_n + b_i, \sigma_i) \mathcal{N}(\mathbf{x}_n;\mathbf{c}_i, \bm{\Gamma}_i)}.
\end{align}
Maximizing (\ref{eq:Q_function}) with respect to each of the model parameters in \begin{math}\theta\end{math} we obtain the parameter-update equations below:
\begin{align}
\label{eq:Max_c}
\mathbf{c}_k &= \sum_{n=1}^{N}\frac{r_{nk}}{r_{k}}\mathbf{x}_n,\\
\label{eq:Max_Gamma}
\bm{\Gamma}_k &= \sum_{n=1}^{N}\frac{r_{nk}}{r_{k}}(\mathbf{x}_n - \mathbf{c}_k)(\mathbf{x}_n - \mathbf{c}_k)^{\top},\\
\label{eq:Max_A}
\mathbf{A}_k &= \overline{Y}_k\overline{\mathbf{X}}_k^{\dagger},\\
\label{eq:Max_b}
\mathbf{b}_k &= \sum_{n=1}^{N}\frac{r_{nk}}{r_{k}}(y_n - \mathbf{A}_k\mathbf{x}_n),\\
\label{eq:Max_sigma}
\sigma_k &= \sum_{n=1}^{N}\frac{r_{nk}}{r_{k}}(y_n - \mathbf{A}_k \mathbf{x}_n - b_k)^2,\\
\label{eq:Max_pi}
\pi_k &= r_k / \sum_{k=1}^{K}\nolimits r_k,
\end{align}
where \begin{math}\dagger\end{math} is the Moore-Penrose pseudo inverse operator and
\begin{math}
\overline{\mathbf{X}}_k = \{ \sqrt{r_{nk}}(\mathbf{x}_n-\mathbf{c}_k)\}_{n=1}^{N},
\overline{Y}_k = \{ \sqrt{r_{nk}}(y_n-\bar{y}_k)\}_{n=1}^{N}
\end{math}
are sets of centred and weighted points with
\begin{math}
\bar{y}_k = \sum_{n=1}^{N}\nolimits\frac{r_{nk}}{r_{k}} y_n
\end{math}.
The RPL-GMR algorithm should be evaluated until convergence of the expected complete-data log-likelihood in (\ref{eq:Q_function}). A typical convergence criterion might be
\begin{align}
\mathcal{L}(\theta^{(i+1)}) - \mathcal{L}(\theta^{(i)}) < \epsilon\left|\mathcal{L}(\theta^{(i)})\right|,
\end{align}
where $(i)$ denotes the iteration index and \begin{math}\epsilon\end{math} is some constant to be specified.

\section{Implementation details}\label{sec:robust_implementation}

We now describe in detail the implementation of the proposed method. A formal description is provided in Algorithm~\ref{alg:GLLiM_EM} and the effect of each of the stages can be seen in Fig. \ref{fig:Pipeline}.
%
%

\subsection{Initialisation}\label{sec:initialisation}

The RPL-GMR algorithm (as with any EM algorithm) does not necessarily find globally optimal solutions and is therefore sensitive to initial conditions. 
An important aspect of initialisation is the decision of how big a model to use in terms of the number of components. 
\addnote[modelselection]{1}{
There is a general consensus that a computationally efficient and well-founded strategy for mixture-model-selection is to start with an over-estimated number of components and to merge them according to criteria such as minimum message length (MML) \citep{figueiredo2002unsupervised}, Bayes information criterion (BIC) \citep{hennig2010methods}, an entropy criterion  \citep{baudry2010combining}, or measuring pair-wise overlap between components \citep{melnykov2016merging}. We therefore choose to initialise with a large number of components that is likely to be higher than the number of planes we expect to find, relying on our fusing procedure to later reduce the number of components where necessary. By initialising with a relatively large number of components, it also becomes more likely that smaller planes will be captured.
}
\begin{algorithm}[t!]
\caption{RPL-GMR}\label{alg:GLLiM_EM}
\begin{algorithmic}[1]
\Procedure{RPL-GMR}{$\mathbf{X}, Y, K, \epsilon$}
\State $\theta^{(i)} = KMeansInit(\mathbf{X}, Y, K)$
\State $\mathcal{L}(\theta^{(i)}) = 0$

\State $\rhd$ INITIAL EXPECTATION STEP
\State $u_{nk}^{(i)} = p(y_n|\mathbf{x}_n, Z=k; \theta^{(i)})p(\mathbf{x}_n|Z=k; \theta^{(i)})$
\State $r_{nk}^{(i)} = \pi_k^{(i)} u_{nk}^{(i)}$
\State $r_n^{(i)} = \sum_{k=1}^K r_{nk}^{(i)}$

\Repeat
\State $r_{nk}^{(i)} = r_{nk}^{(i)}/r_{n}^{(i)}$ \Comment{Normalise}
\State $\rhd$ TRIMMING STEP
\State $(\mathbf{X}', Y') = trimming(\mathbf{X}, Y, r_{nk}^{(i)}, u_{nk}^{(i)}, \mathcal{L}(\theta^{(i)}))$
\State $r_k^{(i)} = \sum_{n'=1}^{N'} r_{n'k}^{(i)}$
\State $\pi_k^{(i)} = r_{k}^{(i)}/ \sum_{k=1}^K r_{k}^{(i)}$
\State $\rhd$ MAXIMISATION STEP (using [$\mathbf{X}', Y'$])
\State Compute new $\theta^{(i+1)}$ from equations (\ref{eq:Max_c})-(\ref{eq:Max_pi})
\State $\rhd$ EXPECTATION STEP (using [$\mathbf{X}, Y$])
\State $u_{nk}^{(i+1)} = p(y_n|\mathbf{x}_n, Z=k; \theta^{(i+1)})p(\mathbf{x}_n|Z=k; \theta^{(i+1)})$
\State $r_{nk}^{(i+1)} = \pi_k^{(i+1)} u_{nk}^{(i+1)}$ \Comment{Don't normalise yet}
\State $r_n^{(i+1)} = \sum_{k=1}^K r_{nk}^{(i+1)}$
\State $\mathcal{L}(\theta^{(i+1)}) = \sum_{n=1}^N \log(r_n^{(i+1)})$
\Until {$\mathcal{L}(\theta^{(i+1)}) - \mathcal{L}(\theta^{(i)}) < \epsilon |\mathcal{L}(\theta^{(i)})|$}
\State $\rhd$ POST-PROCESSING
\State $r_{nk}^{(i+1)} = r_{nk}^{(i+1)}/r_{n}^{(i+1)}$ \Comment{Normalise}
\State $r_k^{(i+1)} = \sum_{n'=1}^{N'} r_{n'k}^{(i+1)}$
\State $\pi_k^{(i+1)} = r_{k}^{(i+1)}/ \sum_{k=1}^K r_{k}^{(i+1)}$
\State $clustering_{n'} = max_k(r_{n'k}^{(i+1)})$
\State $\pi_k^{(i+1)} = densityCheck(\mathbf{X}', clustering_{n'}, \pi_k^{(i+1)})$
\State $(\theta^{(i+1)}, X_{seg}) = FuseComponents(\theta^{(i+1)}, r_k^{(i+1)}, \mathbf{X},  Y)$\\
\Return $(\theta^{(i+1)}, X_{seg})$
\EndProcedure
\end{algorithmic}
\end{algorithm}

Plane-size is also an important consideration when deciding on the number of model components for the following reason: Whereas errors in the positions of points \textit{across} planes may obey something like Gaussian distributions, the positions of points \textit{along} planes have distributions that are more uniform in nature. Non-Gaussian distributions can be better described by multiple Gaussians. As a result, our Gaussian components often \textit{prefer} to co-locate, sharing points belonging to a single plane, rather than forcing each other to occupy different planes. With components not always readily re-distributing to other regions, it is important that components are placed with good proximity to all planes during initialisation. For this reason, if a model with too few components is used, data-points belonging to smaller planes will often be neglected. Choosing a relatively large number of initial model components is one way to ensure that smaller planes are also captured. On the other hand, fitting too many model components is computationally expensive and can lead to over-fitting where components fit to noise, ignoring larger-scale patterns in the data. The choice of the number of model components is therefore data-dependent and is a hyper-parameter that must be tuned.

\begin{figure*}[t]
\centering
\captionsetup{justification=centering}
     \begin{subfigure}[b]{0.329\linewidth}
         \centering
         \includegraphics[width=0.95\linewidth]{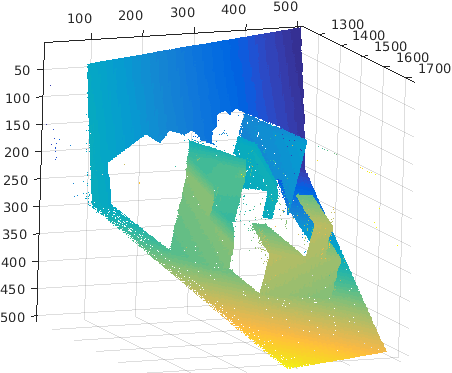}
         \caption{3D point-cloud input.}
         \label{fig:uv1z}
     \end{subfigure}%
     \begin{subfigure}[b]{0.329\linewidth}
         \centering
         \includegraphics[width=0.95\linewidth]{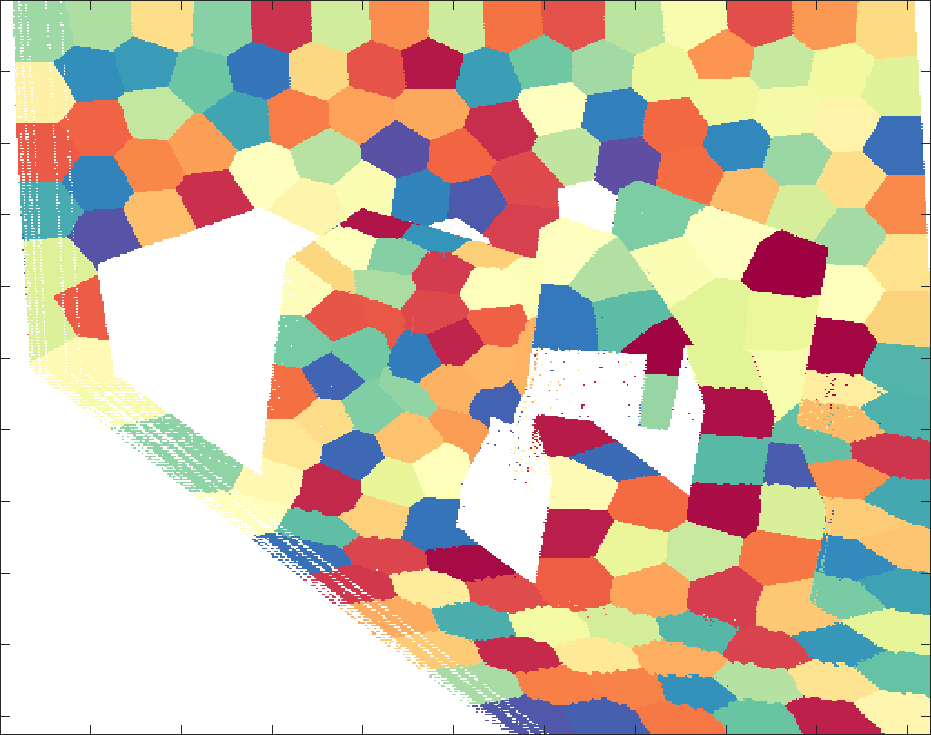}
         \caption{K-means initialisation.}
         \label{fig:init}
     \end{subfigure}
     \begin{subfigure}[b]{0.329\linewidth}
         \centering
         \includegraphics[width=0.95\linewidth]{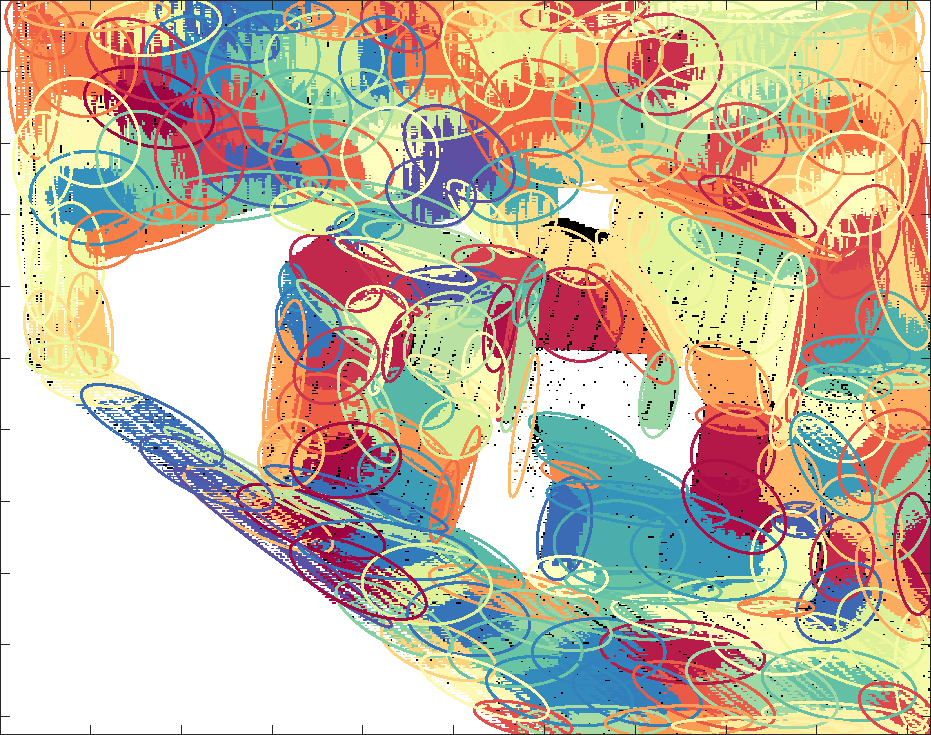}
         \caption{Clusters based on MAP}
         \label{fig:IT_100_gauss}
     \end{subfigure}
     \\
     \begin{subfigure}[b]{0.329\linewidth}
         \centering
         \includegraphics[width=0.95\linewidth]{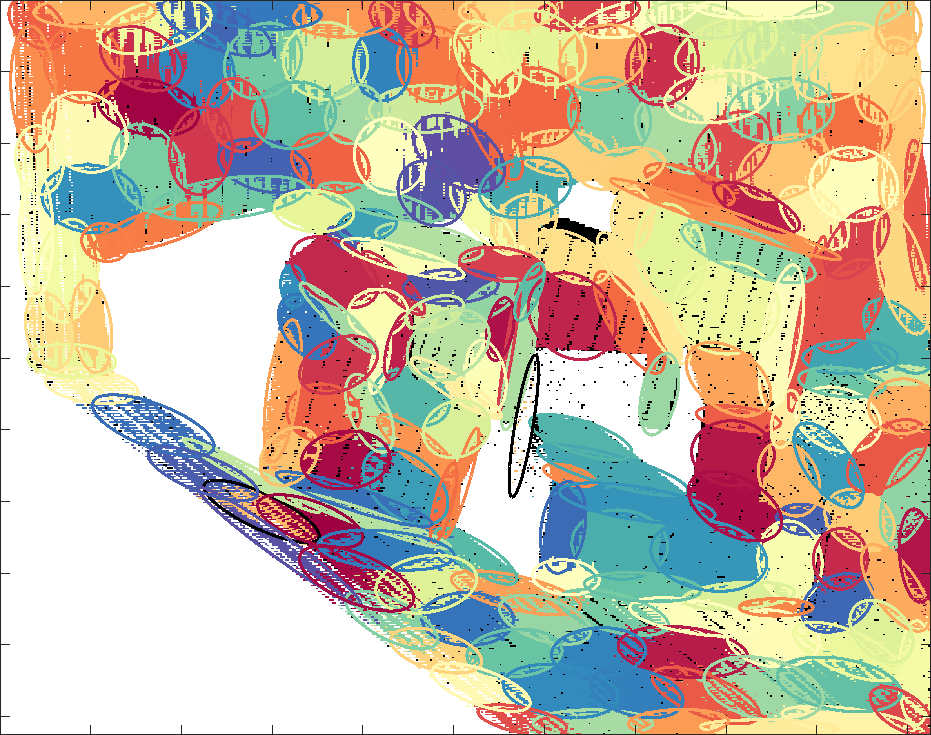}
         \caption{Outlying components detected}
         \label{fig:density_check}
     \end{subfigure}
     \begin{subfigure}[b]{0.329\linewidth}
         \centering
         \includegraphics[width=0.95\linewidth]{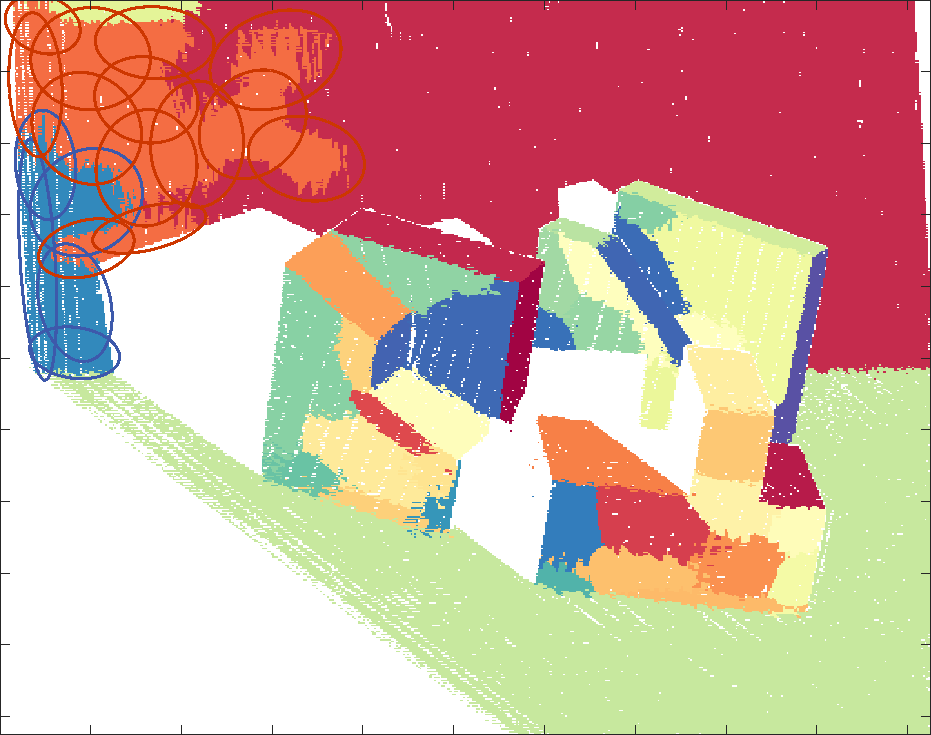}
         \caption{Intermediate fusion.}
         \label{fig:Fusing_intermediate}
     \end{subfigure}
     \begin{subfigure}[b]{0.329\linewidth}
         \centering
         \includegraphics[width=0.95\linewidth]{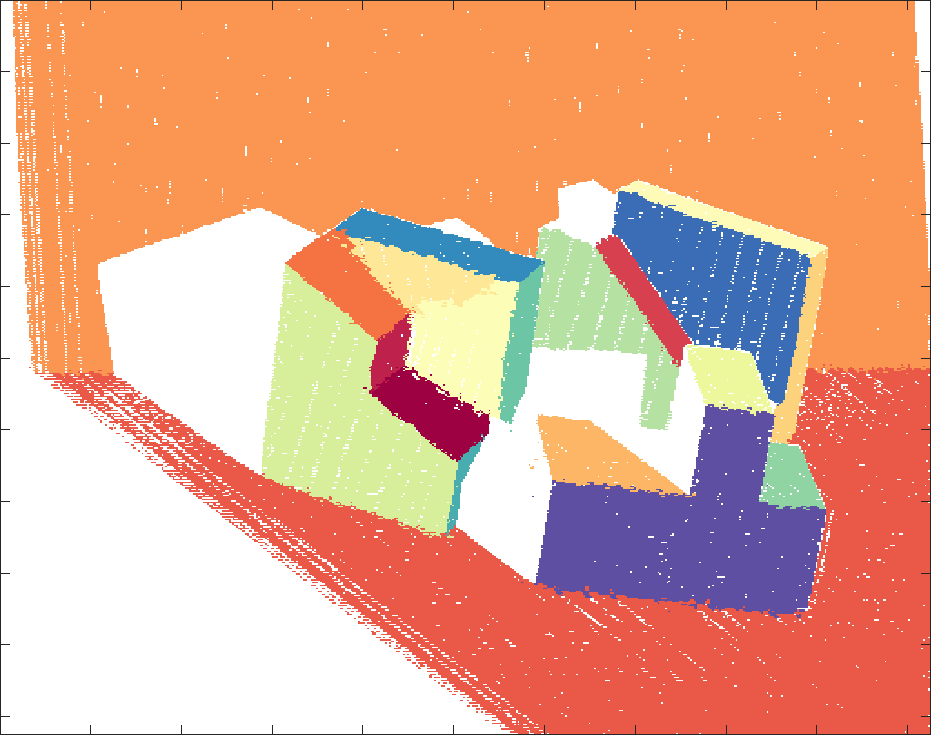}
         \caption{Final result}
         \label{fig:Fusing_output}
     \end{subfigure}
\caption{Visualisation of various stages of Algorithm \ref{alg:GLLiM_EM}.} \label{fig:Pipeline}
\end{figure*}

Initial model parameters are calculated from clusters found by applying randomly initialised k-means to the 3D point set.
An example of output of this initialisation procedure is shown in Fig.~\ref{fig:init}. Also tested was initialisation using points within squares of a regular grid. RPL-GMR was found to converge more quickly when initialised with k-means than with the regular grid; perhaps because, despite not \textit{knowing} about planes in the data, k-means is still able to capture edges where one plane occludes another and there is a large difference in the proximity of points between planes. 

\subsection{Robustness to outliers}\label{sec:robust_EM}

Data-sets containing outliers can introduce biases into model parameters during plane-fitting and can even lead to completely spurious planes being found. As mentioned in Section \ref{sec:related_work}, many plane-extraction methods achieve robustness by fitting planes using RANSAC. Instead, we embed a \textit{trimming} step, inspired by \citep{c12}, within the body of the EM procedure: At each iteration of RPL-GMR, points are ranked based on how likely they are to be outliers, then a certain fraction are discarded (or \textit{trimmed}) from the top of the ranking before continuing to the maximisation step. The general assumption made during trimming is that the parameters of the model are initialised (and remain) to be close to their ideal values. If this is true then outliers can be identified based on their agreement (or lack of agreement) with the current estimate of the model.

To trim successfully, we need two things: (i) reasonable knowledge of the number of outlying data-points; and (ii) a score by which the data-points can be ranked in order of likelihood that they are outliers. Knowledge of the number of outlying data-points can be obtained from training data or else from known camera-characteristics. It is better to over-estimate this fraction \citep{c12}. As for the score, \citep{c12} recommends that, for unbalanced Gaussian mixtures, i.e. mixtures where components represent different numbers of data-points, component-wise confidence-level ordering based on Mahalanobis distance from the most likely component-centre should be used. This avoids the trimming of all points belonging to weak components, as might occur with ordering based on posterior probabilities. Posterior probabilities \textit{are} used, however, to associate points with most likely components.
Rather than ordering based on Mahalanobis distances, we use the likelihood, $p(\mathbf{x},y|Z=k;\theta)$, since it is already calculated prior to the trimming step ($u_{nk}$ in Algorithm \ref{alg:GLLiM_EM}). This ordering is equivalent to ordering by Mahalanobis distances since, for Gaussians, Mahalanobis distance is proportional to \begin{math}\sqrt{-\log p(\mathbf{x},y|Z=k;\theta)}\end{math}, and \begin{math}\sqrt{-\log(x)}\end{math} decreases monotonically between 0 and 1.

EM guarantees to increase (\ref{eq:Q_function}) at each iteration. However, by including the trimming step, the data-set used during the maximisation step will likely change at each iteration. This breaks the guarantee of an ever-increasing log-likelihood. 
\addnote[outliers]{1}{
To avoid this problem, additional individual points are trimmed from the sum until the log-likelihood becomes larger than that following the previous iteration. This allows the log-likelihood to be used to test for convergence, even after having removed at least a fraction, $1-\alpha$, of the points}. The maximisation step then improves the parameters to further increase the log-likelihood.

An example of output from RPL-GMR is given in Fig. \ref{fig:IT_100_gauss}. For visualisation purposes, clusterings of points based on MAP are represented by different colours.
In the same colours, we have also plotted contours of constant probability for each of the $\mathcal{X}$-space Gaussians given by equation (\ref{eq:locality}). 
Points that have been trimmed are shown in black, including those of a plane for which a component was unfortunately not found. 
\addnote[density_check]{1}{
In Fig.~\ref{fig:Pipeline}d, two deleted components (black contours) are shown. These were removed by an additional \textit{densityCheck()} function (see Algorithm~\ref{alg:GLLiM_EM}) that attempts to detect and remove any outlying components (components fitted only to outlying data-points) by comparing the density of MAP-clustered points in $\mathcal{X}$-space to a threshold, $T_{\rho}$.
}
\begin{table*}[!ht]
\caption{\label{tab:SegCompABW}
\textit{SegComp} benchmarking results using the test data of the ABW and PERCEPTRON datasets. The best results are shown in \textbf{bold} and the second-best results are shown in \textit{\textbf{slanted bold}}. Our method yields very good results (second best and third best in terms of number of correctly detected planes. Overall, RPL-GMR is the second best performing method.}
\begin{center}
\resizebox{\textwidth}{!}{%
\begin{tabular}{|c|c|c|c|c|c|c|}
\hline
\textbf{Method} & \textbf{Correctly detected} & \textbf{Orientation deviation} & \textbf{Over-seg.} & \textbf{Under-seg.} & \textbf{Missed} & \textbf{Spurious} \\
\hline
\hline
\multicolumn{7}{|c|}{\textit{SegComp} ABW data-set (30 test images) \citep{c18}. Scores calculated using a threshold of 80\% pixel-overlap.} \\ \hline
USF \citep{c10_16} & 12.7 / 15.2 (83.5\%) & 1.6 & \textit{\textbf{0.2}} & \textit{\textbf{0.1}} & 2.1 & 1.2 \\ \hline
WSU \citep{c10_16} & 9.7 / 15.2 (63.8\%) & 1.6 & 0.5 & 0.2 & 4.5 & 2.2 \\ \hline
UB \citep{c10_16} & 12.8 / 15.2 (84.2\%) & \textbf{1.3} & 0.5 & \textit{\textbf{0.1}} & 1.7 & 2.1 \\ \hline
UE \citep{c10_16} & \textbf{13.4 / 15.2 (88.1\%)} & 1.6 & 0.4 & 0.2 & \textit{\textbf{1.1}} & \textit{\textbf{0.8}} \\ \hline
UFPR \citep{c10_16} & 13.0 / 15.2 (85.5\%) & 1.5 & 0.5 & 0.1 & 1.6 & 1.4 \\ \hline
Oehler et al. \citep{c9_29} & 11.1 / 15.2 (73.0\%) & \textit{\textbf{1.4}} & \textit{\textbf{0.2}} & 0.7 & 2.2 & \textit{\textbf{0.8}} \\ \hline
Holz et al. \citep{c5_19} & 12.2 / 15.2 (80.1\%) & 1.9 & 1.8 & \textit{\textbf{0.1}} & \textbf{0.9} & 1.3 \\ \hline
Feng et al. \citep{c7_11} & 12.8 / 15.2 (84.2\%) & 1.7 & \textbf{0.1} & \textbf{0.0} & 2.4 & \textbf{0.7} \\ \hline
\textit{\textbf{RPL-GMR}} (proposed) & \textit{\textbf{13.1 / 15.2 (85.8\%)}} & 1.6 & \textit{\textbf{0.2}} & \textit{\textbf{0.1}} & 1.8 & \textit{\textbf{0.8}} \\ \hline
\hline
\multicolumn{7}{|c|}{\textit{SegComp} PERCEPTRON data-set (30 test images) \citep{c18}. Scores calculated using a threshold of 80\% pixel-overlap.} \\ \hline
USF \citep{c10_16} & 8.9 / 14.6 (60.9\%) & 2.7 & 0.4 & \textbf{0.0} & 5.3 & 3.6 \\ \hline
WSU \citep{c10_16} & 5.9 / 14.6 (40.4\%) & 3.3 & 0.5 & 0.6 & 6.7 & 4.8 \\ \hline
UB \citep{c10_16} & 9.6 / 14.6 (65.7\%) & 3.1 & 0.6 & \textit{\textbf{0.1}} & 4.2 & 2.8 \\ \hline
UE \citep{c10_16} & 10.0 / 14.6 (68.4\%) & 2.6 & \textbf{0.2} & 0.3 & 3.8 & 2.1 \\ \hline
UFPR \citep{c10_16} & \textbf{11.0 / 14.6 (75.3\%)} & \textit{\textbf{2.5}} & \textit{\textbf{0.3}} & \textit{\textbf{0.1}} & 3.0 & 2.5 \\ \hline
Oehler et al. \citep{c9_29} & 7.4 / 14.6 (50.1\%) & 5.2 & \textit{\textbf{0.3}} & 0.4 & 6.2 & 3.9 \\ \hline
Holz et al. \citep{c5_19} & \textbf{11.0 / 14.6 (75.3\%)} & 2.6 & 0.4 & 0.2 & \textbf{2.7} & \textbf{0.3} \\ \hline
Feng et al. \citep{c7_11} & 8.9 / 14.6 (60.9\%) & \textbf{2.4} & \textbf{0.2} & 0.2 & 5.1 & 2.1 \\ \hline
\textit{\textbf{RPL-GMR}} (proposed) & \textit{\textbf{10.6 / 14.6 (72.4\%)}} & \textit{\textbf{2.5}} & \textit{\textbf{0.3}} & 0.3 & \textit{\textbf{3.0}} & \textit{\textbf{2.0}} \\ \hline
\end{tabular}}
\end{center}
\end{table*}

\subsection{Fusing of planar Gaussian components}\label{sec:Fusing}

At first glance, rather than combining components as a post-processing stage, it might seem that it would be more elegant to include some form of model-selection within the RPL-GMR loop. In \citep{figueiredo2002unsupervised}, for example, the number of model components is gradually reduced during the EM procedure until the most parsimonious description of the data is found, as measured by a Minimum Message Length (MML) criterion. In our case, however, reducing the number of components based on MML is not meaningful since our distributions of points are non-Gaussian. E.g. many of the planar surfaces are rectangular and are more effectively modelled by multiple components. Rather than reducing the number of components \textit{during} EM iterations, our approach is, instead, to fuse components together as a post-processing stage. By doing so, we are able to obtain more accurate estimates of the plane parameters by combining information from multiple co-planar components, but are also able to maintain the associations of data points with the original set of model components since no further expectation steps are performed following the fusing stage.

Components are fused together if three criteria are met: 1) The components must be adjacent to one another; 2) By combining the components, the RMS probability-weighted deviation of points perpendicular to the combined plane must not exceed a certain threshold; and 3) Each of the components being fused must not protrude too far from the plane of the other component. Similar to \citep{c7_11}, we first build an adjacency graph of clusters in the 2D $\mathcal{X}$-space. In \citep{c7_11} this is straightforward as their data are divided into a regular grid. In our case, model components are scattered throughout the data and we must explicitly test for adjacency. To do this, we test for overlap of ellipses formed from the Mahalanobis distances of the Gaussians in (\ref{eq:locality}), scaled by a factor $c_{D_{M}}$. An efficient method for testing the overlap of ellipses can be found in \citep{c17}. An alternative approach could be to test for adjacency of convex cells in the 3D Voronoi tessellation formed by MAP partitioning of the space about the mixture model.

In order to test the second and third fusing-criteria, we make use of the principal components of variation in the data as weighted by the responsibilities found for each component. I.e. for each component, we calculate eigenvalues of the responsibility-weighted data: the smallest eigenvalue is equivalent to the mean squared error (MSE) of points from the plane. 
In practice we calculate the eigenvalues of matrix (\ref{eq:joint_mean_and_cov}).

The fusing algorithm proceeds as follows: The node in the adjacency graph whose component has the smallest MSE is identified; \textit{hypothetical} combinations are then made with each adjacent component to find the plane with the lowest combined MSE. No combination is made if the best resulting MSE is greater than a certain threshold, \begin{math}T_{MSE}\end{math}, or if the third fusing-criterion is not met (discussed below). Fusing will terminate once each combination of adjacent nodes has been tested. In \citep{c7_11}, the MSEs are stored in a min-heap data-structure for efficiency. This could be done here as well. However, the cost of running our fusing algorithm is already much less than running RPL-GMR.

Up to this point in the algorithm, we have made efforts to ensure that smaller planes in our unbalanced mixture are not lost. For example, one reason we initialise with a large number of components is to capture smaller planes. We also used component-wise confidence-level ordering during trimming to avoid the loss of smaller planes. Without the third fusing-criterion, however, smaller planes could easily be subsumed by larger ones, provided the MSE remains low enough. In some cases the data-based distance metric of combined MSE works well. E.g. dominant planes are able to \textit{mop up} small erroneous planar components fitted to noise at the edges of true clusters, despite having orientations roughly perpendicular to the main plane. If the smaller plane extends significantly beyond the noise of the more dominant plane, however, then we probably don't want to merge the two. Before merging any two components, therefore, we perform the third check on the magnitudes of projections of the two main eigenvectors (in both negative and positive directions) onto the other plane's normal. The test fails if, for both planes, the magnitude of any of these four projections is greater than a certain threshold: $T_{proj} \times \sqrt{\textrm{MSE}}$.

\section{Benchmarking}\label{sec:benchmarking}

We evaluated the RPL-GMR algorithm using the \textit{ABW} and \textit{PERCEPTRON} data-sets available as part of the \textit{SegComp} (Segmentation Comparison) project from the University of South Florida \citep{c18}. Both of these data-sets contain depth-images of entirely planar scenes along with ground-truth segmentations. The images of the ABW data-set were taken using an ABW structured light camera whereas the PERCEPTRON camera uses scanning laser range finding (LRF) technology. Each set contains 10 training-images and 30 test-images. The \textit{SegComp} package also includes an automated comparison program that compares segmented images with the ground-truth segmentations and produces various statistics. As well as comparing clustered pixels in the image, the program compares the orientations of the model planes that were found.

\begin{figure*}[t]
\centering
\captionsetup{justification=centering}
         \begin{subfigure}[b]{0.193\textwidth}
                 \centering
                 \includegraphics[width=1.0\textwidth]{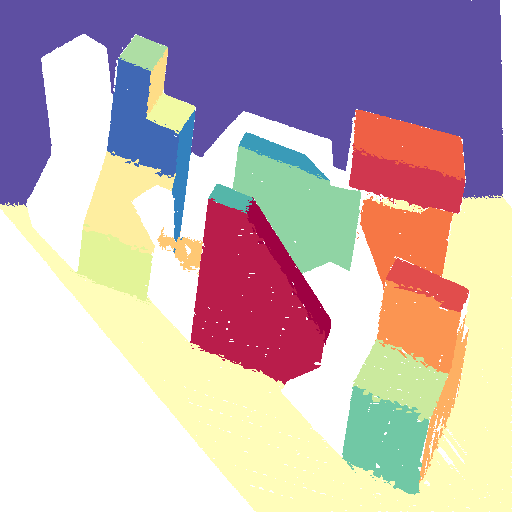}
                 \caption{21/27, 6 missed, 3 spurious}
                 \label{fig:abw.0}
         \end{subfigure}
         \begin{subfigure}[b]{0.193\textwidth}
                 \centering
                 \includegraphics[width=1.0\textwidth]{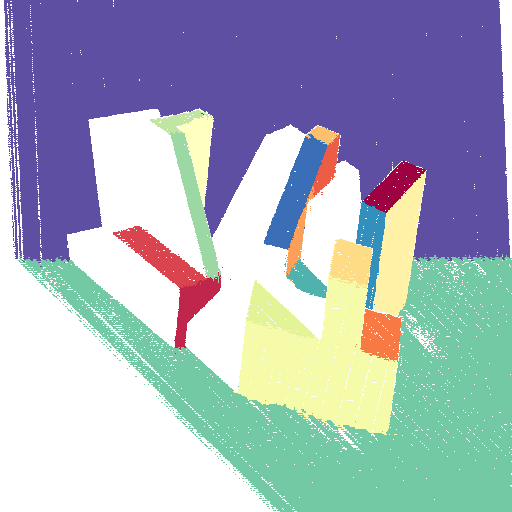}
                 \caption{16/19, 1 over, 2 missed, 1 spurious}
                 \label{fig:abw.4}
         \end{subfigure}%
         \begin{subfigure}[b]{0.193\textwidth}
                 \centering
                 \includegraphics[width=1.0\textwidth]{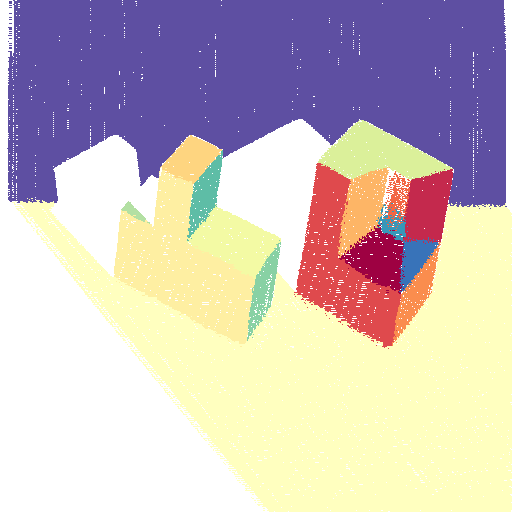}
                 \caption{14/17, 3 missed, 3 spurious}
                 \label{fig:abw.9}
         \end{subfigure}
         \begin{subfigure}[b]{0.193\textwidth}
                 \centering
                 \includegraphics[width=1.0\textwidth]{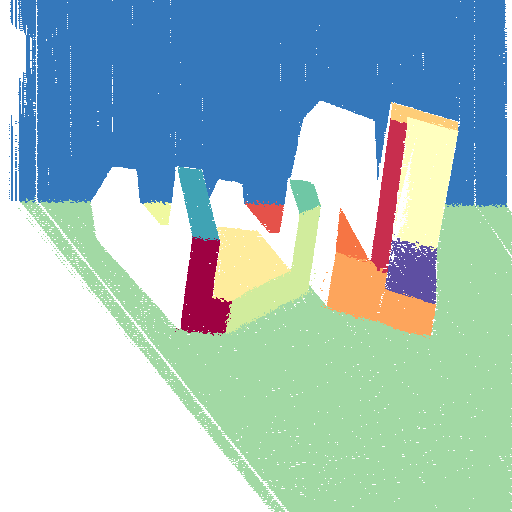}
                 \caption{15/17, 2 missed \\ \quad}
                 \label{fig:abw.14}
         \end{subfigure}
         \begin{subfigure}[b]{0.193\textwidth}
                 \centering
                 \includegraphics[width=1.0\textwidth]{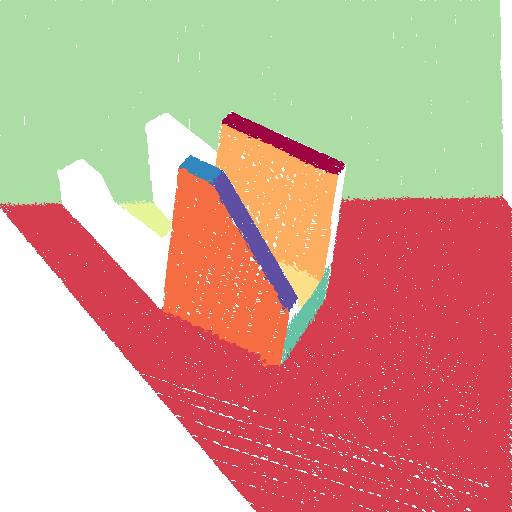}
                 \caption{9/10, 1 missed, 1 spurious}
                 \label{fig:abw.21}
         \end{subfigure}

     \begin{subfigure}[b]{0.193\textwidth}
             \centering
             \includegraphics[width=1.0\textwidth]{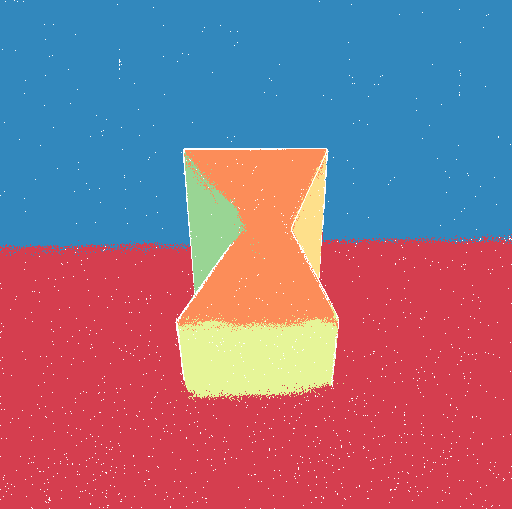}
             \caption{6/6 \\ \quad}
             \label{fig:perc.1}
     \end{subfigure}
     \begin{subfigure}[b]{0.193\textwidth}
             \centering
             \includegraphics[width=1.0\textwidth]{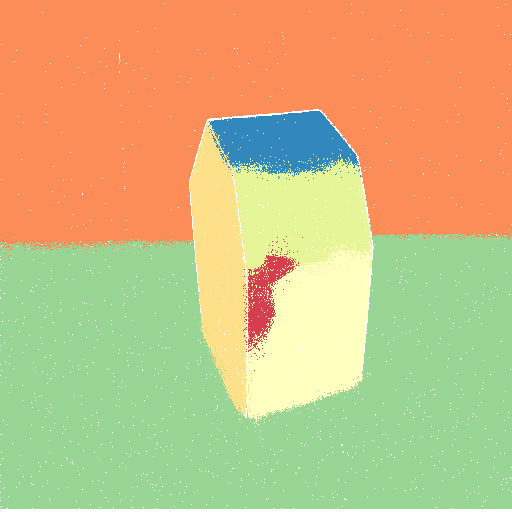}
             \caption{6/6, 1 spurious \\ \quad}
             \label{fig:perc.5}
     \end{subfigure}
     \begin{subfigure}[b]{0.193\textwidth}
             \centering
             \includegraphics[width=1.0\textwidth]{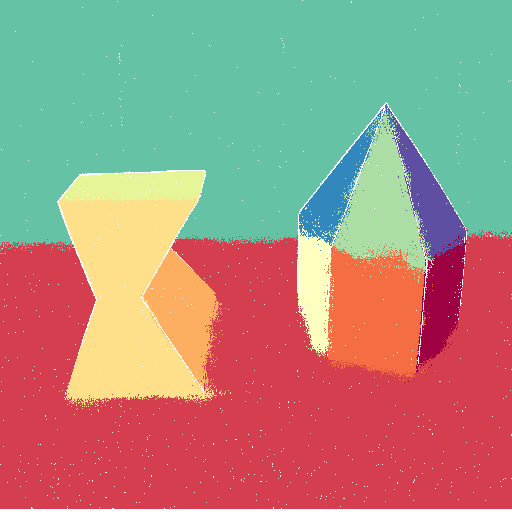}
             \caption{11/11 \\ \quad}
             \label{fig:perc.16}
     \end{subfigure}
     \begin{subfigure}[b]{0.193\textwidth}
             \centering
             \includegraphics[width=1.0\textwidth]{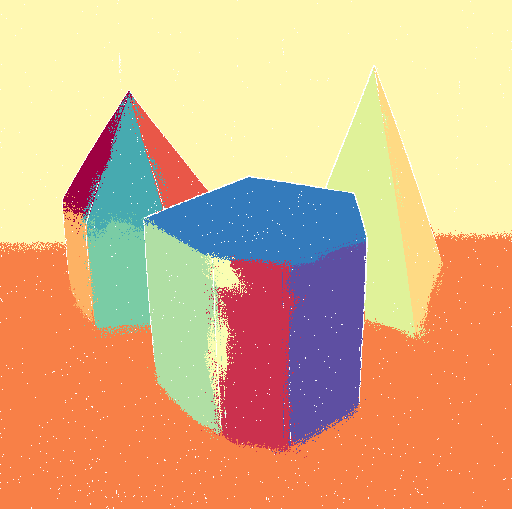}
             \caption{13/13, 1 spurious \\ \quad}
             \label{fig:perc.24}
     \end{subfigure}
     \begin{subfigure}[b]{0.193\textwidth}
             \centering
             \includegraphics[width=1.0\textwidth]{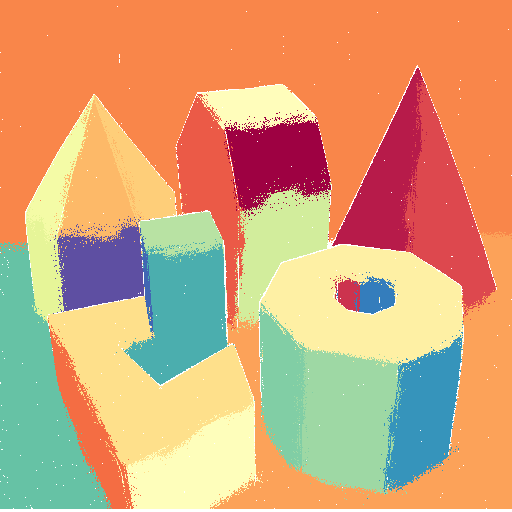}
             \caption{22/30, 2 under, 4 missed, 2 spurious}
             \label{fig:perc.29}
     \end{subfigure}
\caption{Results from the \textit{SegComp} benchmark using RPL-GMR. Examples from ABW test data (top row) and from the PERCEPTRON test data (bottom row).} \label{fig:ms-seg_files}
\end{figure*}

When working with depth-images, it is advantageous to be able to use image-coordinates as values in the $\mathcal{X}$-space of the RPL-GMR model. Doing so avoids potential problems with the degeneracy of points that happen to share the same xy-coordinates in Cartesian space. (In an image, each point has its own, non-degenerate uv-coordinate.) The transformation from image-space to depths, however, is nonlinear. To avoid this problem, it is necessary to work with a quantity that is inversely proportional to depth, such as disparity. In our evaluation we worked with the quantity $s/Z$ using the scale-factor $s = \frac{(\text{rows}+\text{cols})/2}{(1/z)_{max} - (1/z)_{min}}$. Without the scale-factor, inverse depths (which tend to be very small values) have little effect during the EM procedure and the algorithm struggles to differentiate between nearby planes of different orientations. $U$, $V$ and $s/Z$ are the axes plotted in Fig. \ref{fig:uv1z}.

\addnote[parameters]{1}{
The parameters of our algorithm were tuned by experimenting on the ABW and PERCEPTRON training sets. We arrived at the following settings: $K = 200$ (the number of model components) was chosen as it gives an initial clustering similar in size to the smallest planes of the training sets, e.g. Fig.~\ref{fig:Pipeline}b; $c_{D_M} = 2.1$ (the ellipse size used for adjacency checking of components) was chosen such that the ellipse sizes, e.g. Fig.~\ref{fig:Pipeline}c, roughly contain all MAP-clustered points; $T_{proj} = 10$ (the parameter that avoids fusing of small planar components with large perpendicular planes) was chosen so that the vector $(T_{proj} \: \sqrt{MSE} \:) \mathbf{v}$, where $\mathbf{v}$ is the unit vector normal to the plane, would not extend too far beyond the cloud of noisy points belonging to that plane; the simple value of $T_{\rho} = 0.5$ (the threshold for the outlying-component-check, based on density in $\mathcal{X}$-space) was chosen to signify that, using Fig.~\ref{fig:Pipeline}d as reference, at least 50\% of the pixels contained within any ellipse must be MAP-associated with that component in order to be considered valid.
}

\addnote[parameters-more]{1}{
For $\alpha$, values were taken directly from the provided ground-truth segmentations of the training sets ($\alpha=0.98$ for ABW and $\alpha=0.99$ for PERCEPTRON). However, as previously noted, using an over-estimate of $\alpha=0.98$ for both data-sets would also have been appropriate. To tune $T_{MSE}$, crude parameter searches were performed in the range $[1.5, 12]$ at intervals of $0.5$. This arrived at values of $T_{MSE} = 5$ for ABW and $7.5$ for PERCEPTRON. However, we noticed that above a value of around 5, sensitivity to the parameter was low due to the additional test on $T_{proj}$ before fusing components. Note that $T_{proj}$ requires less tuning as the threshold used is also proportional to $\sqrt{MSE}$. We did not tune the stopping criterion of the EM algorithm and set this to the relatively strict value of $\epsilon = 10^{-5}$, which was never achieved during benchmarking. In all cases we stopped the EM algorithm after a maximum of 50 iterations, which was enough to reach a satisfactory level of convergence and produce the near-optimal results. The automated results of running on the \textit{SegComp} test sets are given in Table~\ref{tab:SegCompABW} and a selection of images for direct comparison with those shown in \citep{c7_11}  are shown in Fig.~\ref{fig:ms-seg_files}.
}


Comparison of results in Table \ref{tab:SegCompABW} shows that RPL-GMR performs consistently well by most of the measures. Out of the nine methods, for the ABW data-set, RPL-GMR ranks second (or joint-second) for four out of the six measures  (correct detections, over-segmentation, under-segmentation, and for not producing spurious planes). For the orientation-deviation and missed planes metrics, RPL-GMR ranks lower: joint-fourth and fifth, respectively. However, scores for these metrics are well within the normal range. For the PERCEPTRON data-set, RPL-GMR ranks as second for not producing spurious planes, joint-second for both orientation-deviation and for not missing planes, and third and joint-third for correctly detecting planes and not over-segmenting them. RPL-GMR ranked as only joint-sixth for under-segmentation. However, again, this score is well within the normal range.

For qualitative evaluation, a selection of segmented images is displayed in Fig. \ref{fig:ms-seg_files}. 
\addnote[comments]{1}{
These can be compared directly with those presented in \citep{c7_11}. According to Table~\ref{tab:SegCompABW}, our algorithm over- and under-segments slightly more than \citep{c7_11} but correctly detects significantly more planes. The only images in Fig.~\ref{fig:ms-seg_files} that are representative of this happen to be Figures \ref{fig:ms-seg_files}h, \ref{fig:ms-seg_files}i and \ref{fig:ms-seg_files}j from the PERCEPTRON data-set. In general, it seems as though our method was better able to handle the noisier PERCEPTRON data-set than \citep{c7_11}. In Figures \ref{fig:ms-seg_files}h and \ref{fig:ms-seg_files}j we were able to meet the 80\% pixel-overlap threshold for correctly detecting planes whereas \citep{c7_11} seems to have missed planes by discarding many noisy pixels or else over-segmenting due to the noise. In Fig.~\ref{fig:ms-seg_files}j it is more obvious that our method has performed better, correctly capturing three rather subtle planes: one tightly angled plane on the left-hand side of a box (shown in blue), and two small planes at subtly different angles inside the octagonal, toric object. Despite these successes, there remains some room for improvement}.
In Fig.~\ref{fig:abw.21} a large section of a plane has been missed. This seems to have been caused by a combination of unfortunate initialisation and a value of $\alpha$ that was perhaps slightly too small for the image. One solution might be to initialise with a larger number of components, but at greater computational cost. Another problem that can be seen is the under-segmentation of planes in Figures \ref{fig:abw.9}, \ref{fig:perc.5} and \ref{fig:perc.24}. These issues seem to have been misdiagnosed by the automated \textit{SegComp} comparison program as spurious planes since the largest parts of the planes were captured correctly. These problems of over-segmentation could potentially be solved by better tuning of the $T_{proj}$ and $T_{MSE}$ parameters. A coarse hyper-parameter search \textit{was} performed, however.
\section{Conclusions}\label{sec:conclusions}

\addnote[conclusions]{1}{
We have shown that the proposed RPL-GMR algorithm can be used successfully to extract planar patches from depth-data. Combined with an outlier-trimming step embedded within the EM procedure to achieve robustness and with a component-fusing method, benchmark results place our algorithm among the top-performing algorithms in the recent literature in terms of segmentation-quality. The proposed method processes 3D point clouds with no prior information about the sensor being used. RPL-GMR is slower than other recent methods, due to the \textit{batch} nature of EM. However, several strategies could be used to accelerate the algorithm, for example by assuming that the data have a grid-like structure, which enables efficient implementations of region growing, e.g. \citep{c7_11} and \citep{c5_19}. The most time-consuming part of the algorithm 
is the E-step which repeatedly computes the Mahalanobis distance between the cluster centers and all the data points. Several data sampling strategies could be used to speed up the execution of EM, such as running K-means with the desired number of sampled points and then replacing the small clusters of points thus obtained with the cluster centers. One also notices that the proposed algorithm could be used to find an initial segmentation before being applied incrementally as new data become
available, e.g. \citep{evangelidis2017joint}.
}





\section*{Acknowledgments}
Financial support from the European Union via the ERC Advanced Grant \#340113 \textit{Vision and Hearing In Action} (VHIA) is greatly acknowledged.

\bibliographystyle{model2-names}

\end{document}